\documentclass[9pt,twocolumn,twoside]{pnas-new}

\templatetype{pnasmathematics} 
\usepackage{amsthm}

\newcommand{\Continue}{\textbf{continue}}

\newcommand{\sbf}{\mathbf{s}} \newcommand{\tbf}{\mathbf{t}} 
  \newcommand{\xbf}{\mathbf{x}}
\newcommand{\ybf}{\mathbf{y}}

\newcommand{\mubf}{\boldsymbol{\mu}}

\newcommand{\Sigmabf}{\boldsymbol{\Sigma}}
\newcommand{\epsilonbf}{\boldsymbol{\epsilon}}

\newcommand{\Abf}{\mathbf{A}}  \newcommand{\Cbf}{\mathbf{C}}
 \newcommand{\Ebf}{\mathbf{E}} 
  \newcommand{\Ibf}{\mathbf{I}}
  \newcommand{\Lbf}{\mathbf{L}}
\newcommand{\Mbf}{\mathbf{M}}  
  \newcommand{\Rbf}{\mathbf{R}}
\newcommand{\Sbf}{\mathbf{S}}  \newcommand{\Ubf}{\mathbf{U}}
\newcommand{\Vbf}{\mathbf{V}}  \newcommand{\Xbf}{\mathbf{X}}
\newcommand{\Ybf}{\mathbf{Y}}

\newcommand{\zeros}{\textbf{0}}

\newtheorem{Theorem}{Theorem}
\newtheorem{remark}{Remark}
\newtheorem{corollary}{Corollary}

\title{Learning Conserved Networks from  Flows}

\author[a,d]{Satya Jayadev P}
\author[b,d,2]{Shankar Narasimhan} 
\author[c,d,2]{Nirav Bhatt}

\affil[a]{Department of Electrical Engineering, Indian Institute of Technology Madras, Chennai-600036, India}
\affil[b]{Department of Chemical Engineering, Indian Institute of Technology Madras, Chennai-600036, India}
\affil[c]{Department of Biotechnology, Indian Institute of Technology Madras, Chennai-600036, India}
\affil[d]{Robert Bosch Centre for Data Science and Artificial Intelligence, Indian Institute of Technology Madras, Chennai-600036, India}

\leadauthor{Satya Jayadev} 


\begin{abstract}
A challenging problem in complex networks is the network reconstruction problem from data. This work deals with a class of networks denoted as conserved networks, in which a flow associated with every edge and the flows are conserved at all non-source and non-sink nodes. We propose a novel polynomial time algorithm to reconstruct conserved networks from flow data by exploiting graph theoretic properties of conserved networks combined with learning techniques. We prove that exact network reconstruction is possible for arborescence networks. We also extend the methodology for reconstructing networks from noisy data and explore the reconstruction performance on arborescence networks with different structural characteristics. 
\end{abstract}

\dates{This manuscript was compiled on \today}

\begin{document}

\maketitle
\thispagestyle{firststyle}
\ifthenelse{\boolean{shortarticle}}{\ifthenelse{\boolean{singlecolumn}}{\abscontentformatted}{\abscontent}}{}

\dropcap{T}he connectivity between different elements of complex systems in life sciences, engineering, and physics plays an important role in the overall behaviour of these systems \cite{barabasi2016network,boccaletti2006complex}.
Network science deals with characterizing complex networks and understanding their fundamental properties  \cite{barabasi2009scale,watts1998collective,amaral2000classes,nepusz2012controlling}. In network science, such complex systems are represented as graphs where the different elements are modelled as nodes while the interactions between the elements are modelled as edges \cite{barabasi2009scale,nepusz2012controlling}. Techniques from graph theory and dynamical systems have been applied to unravel the organisational principles of the underlying complex processes. \cite{watts1998collective,nepusz2012controlling}.  However, in practice, information related to edge connectivity between the different nodes may not always be available.  In such cases, the edge connectivity information has to be inferred from data. This problem of inferring the edge connectivity from data is one of the important problems in network science and labelled as the network reconstruction problem. 

Several methods based on machine learning and time series analysis in the existing literature have been proposed to infer the underlying network connectivity using steady state or temporal data \cite{han2015robust, angulo2017fundamental,li2017reconstruction, sontag2008network,newman2018network,zavlanos2011inferring} arising in different fields. On the one hand, steady-state data have been used to discover biological networks \cite{sontag2008network}, social science networks \cite{newman2018network} and  engineering networks \cite{pappu2018identifying}. On the other hand, time series data have been used to infer networks in financial networks \cite{materassi2010topological}, biological networks \cite{srividhya2007reconstructing,li2017reconstruction}, and social networks \cite{li2017reconstruction}. 

In general, a network may be used to denote binary relations between entities (eg. social networks).  In addition, the edges may also be directed to indicate the influence direction (eg. causal networks).  In this work, we deal with directed networks, where the direction indicates the transfer of a quantity (flow) from one node to another.  We further assume that the net flow of the quantity at every node is zero, except at source and sink nodes.  In other words, the quantity is conserved at every non-source and non-sink node and there is no loss or accumulation at these nodes. This property of conservation leads to a class of networks which we refer to as \emph{conserved networks} in this work. Flow of metabolic flux from one metabolite to another in biological networks,  flow of commodity (such as power, gas, water) from a sources to consumers in  distribution networks, flow of messages from senders to receivers in communication networks, and money flow in financial networks are examples of conserved networks. 

If the flow data in each edge at different steady states are available, then the question we address is whether it possible to reconstruct the underlying conserved network from flow data? In this work, we show that a conserved network can be reconstructed from flow data by combining properties of the conservation graph and linear learning technique. Specifically, the network reconstruction problem is mapped to the well known graph realization problem by deriving the fundamental cutset matrix of the conservation graph from flow data using principal components analysis. 

The main contributions of this work are as follows: (i) a novel concept of conservation graphs is introduced and their properties related to the network structure are established, (ii)  the properties of conservation graphs are exploited for developing methodology to reconstruct  conserved networks from flow data, (iii)  we show that the developed methodology can reconstruct the underlying conserved network exactly for single-source non-looped networks (or equivalently arborescence networks),  (iv) we extend the proposed methodology to noisy measurements under different scenarios, and  (v) the proposed methodology has a polynomial time complexity. 

\section{Conservation Graphs} \label{Sec:Con_Graph}
A network can be represented as a digraph $G(N,\mathcal{E})$ where $N$ is the set of $|N| = n$ nodes, and $\mathcal{E}$ is the set of $|\mathcal{E}|=e$ directed edges.  Each edge is associated with a flow, and the direction of the edge  indicates the flow direction.  Nodes with no incoming edges are denoted as source nodes and nodes with no outgoing edges are denoted as sink nodes.  All other nodes are denoted as non-sink and non-source nodes and we assume that the network contains $m < n$ such nodes.  Flows associated with edges connected to the sink nodes are designated as sink flows, while all other flows are denoted as non-sink flows. Associated with such a network, we define a conservation graph $G_c(N_c,\mathcal{E})$, with $|N_c| = n_c = m+1$ such that flow conservation equations can be written around every node in the conservation graph $G_c$. It may be noted that although the network is also a graph, it is not a conservation graph because conservation equations cannot be written around source and sink nodes. From a given network topology, the conservation graph is constructed by merging all the source and sink nodes into a single node referred to as the environment node $E$. The edges incident on source and sink nodes are now incident on the environment node $E$ with consistent directions. This ensures that conservation holds at the environment node also.

\begin{figure}[H]
\centering
\includegraphics[scale=0.7]{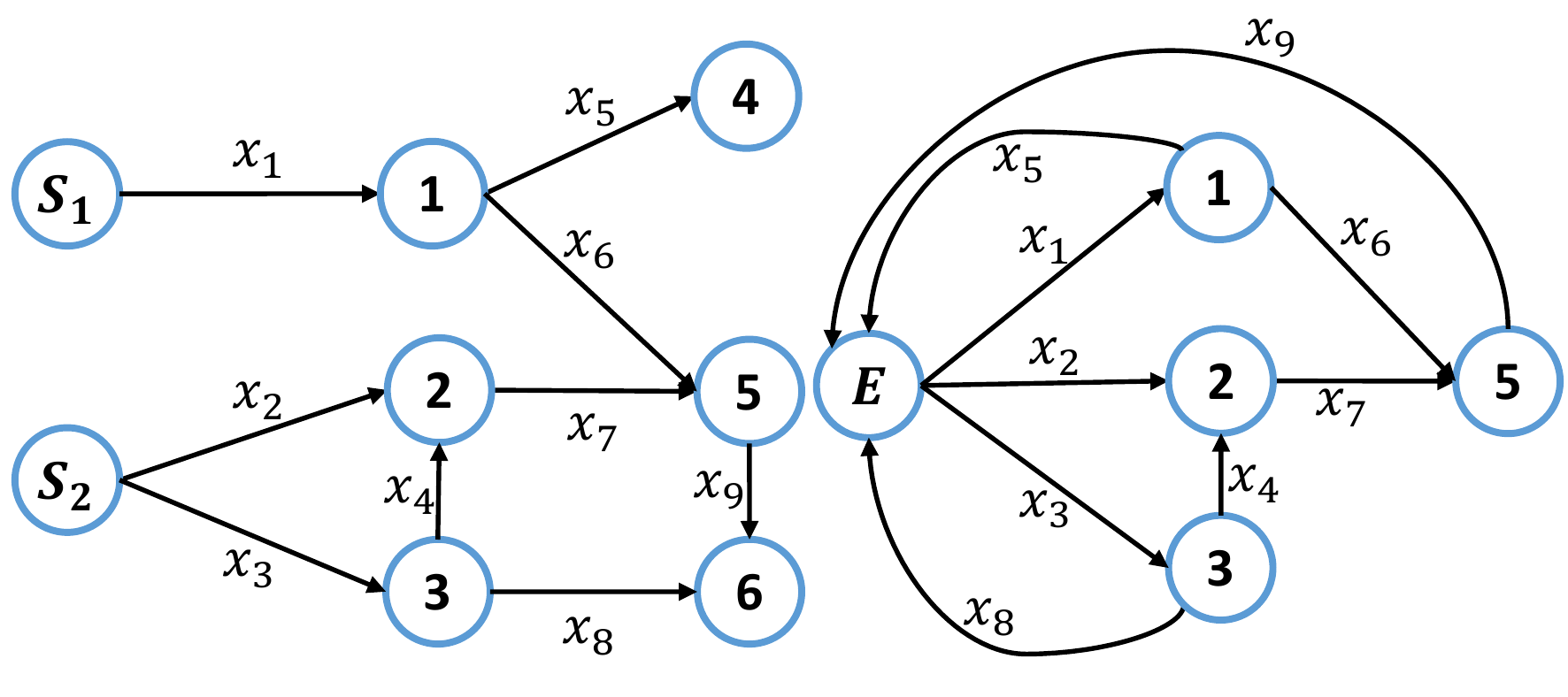} 
\caption{Topology of a network and its conservation graph}
\label{Fig:Network_PNAS}
\end{figure}

For example, consider a simple flow network with two source nodes $S_1$ and $S_2$, two sink nodes $4$ and $6$ and nine edges.  The network is given by graph $G$ in Fig.~\ref{Fig:Network_PNAS}. The conservation graph is constructed by merging all the sources and sinks into a single environment node $E$ as shown in Fig.~\ref{Fig:Network_PNAS}.  Conservation equations can be written around every node of $G_c$ in terms of the steady state flows $x_1,\ldots,x_{9}$ corresponding to the edges \footnote{In this paper, the same labels will be used to indicate the flow variables as well as the edges since there is a unique flow associated with every edge.}. 

The set of independent conservation equations at all nodes can be directly obtained from the $m \times e$ reduced incidence matrix ($\Abf_{ir}$) of $G_c$, whose elements are $0, 1, -1$ depending on the flow directions.  These equations can be written as
\begin{align}\label{Eq:conservation}
\Abf_{ir} \xbf = \zeros,
\end{align}
where $\xbf \in \mathbb{R}^e$ is the vector of flow variables consistent with ordering of edges in $\Abf_{ir}$ and $\zeros$ is a vector of zeros. 

Similarly, we can write another equivalent set of independent conservation equations using the f-cutset matrix with respect to a spanning tree,  $S_{G_c}$, of the conservation graph.  If $\Cbf_f$ is such a f-cutset matrix, we can write:
\begin{align}
\Cbf_f \xbf = \zeros. \label{Eq1}
\end{align}
It is to be noted that $\Abf_{ir}$ and $\Cbf_f$ are linear matrices orthogonal to $\xbf$ and their row spaces are identical. By definition, the rank of both $\Abf_{ir}$ and $\Cbf_f$ is $m$ and is equal to the number of independent equations. For the conservation graph in Fig.~\ref{Fig:Network_PNAS}, a reduced incidence matrix with nodes $1,2,3,5$ on the rows and an f-cutset matrix with respect to spanning tree formed by $x_1,x_2,x_3,x_6$ as branches, are given by:

\begin{align*}
\Abf_{ir} &= 
\begin{array}{c c} 
\begin{array}{c c c c c c c c c}
1 \\
2 \\
3 \\
5
\end{array} 
&
\left[
\begin{array}{c c c c c c c c c}
-1 & 0 & 0 & 0 & 1 & 1 & 0 & 0 & 0\\
0 & -1 & 0 & -1 & 0 & 0 & 1 & 0 & 0\\
0 & 0 & -1 & 1 & 0 & 0 & 0 & 1 & 0 \\
0 & 0 & 0 & 0 & 0 & -1 & -1 & 0 & 1
\end{array}
\right] 
\end{array} \\
\Cbf_f &=
\begin{array}{c c}  &
\begin{array}{c c c c c c c c c} x_1 & x_2 & x_3 & x_6 & x_4 & x_5 & x_7 & x_8 & \; x_9 \\
\end{array} 
\\
&
\left[
\begin{array}{c c c c c c c c c}
\;\; 1 & \;\; 0 & \;\; 0 & \;\; 0 & 0 & -1 & 1 & 0 & -1 \\
\;\; 0 & \;\; 1 & \;\; 0 & \;\; 0 & 1 & 0 & -1 & 0 & 0 \\
\;\; 0 & \;\; 0 & \;\; 1 & \;\; 0 & -1 & 0 & 0 & -1 & 0 \\
\;\; 0 & \;\; 0 & \;\; 0 & \;\; 1 & 0 & 0 & 1 & 0 & -1
\end{array}
\right] 
\end{array}
\end{align*}
In general, the f-cutset matrix with respect to any spanning tree of the conservation graph has the form $[\Ibf \ \ \Cbf_c]$ where the edges corresponding to the columns of the identity matrix are branches and edges corresponding to columns of $\Cbf_c$ are chords of the spanning tree with elements $0, +1, -1$ \cite{Deo}.
In this work, we exploit these conservation linked graph theoretic properties of $G_c$ to identify the network from flow data. 

\section{Problem Formulation and Approach} \label{Sec:Approach}
The objective of this work is to develop a methodology to identify the topology of a network from measurements of flows associated with all its $e$ edges at $n_s > e$ different steady states. The proposed approach to topology identification comprises of two steps: (1) learning the f-cutset matrix of the conservation graph from steady state flow data using principal components analysis (2) realising the conservation graph from the f-cutset matrix. The latter is known as the graph realization problem which has been well studied in the literature \cite{Fuji,Bixby,Parker,Jianping}. It is also known that in general, the graph realized from an f-cutset matrix may not be identical to the original and may be 2-isomorphic to it \cite{whitney1933}.

However, for arborescence networks we prove that it is possible to reconstruct the original network.  We propose a new graph realization algorithm to construct a specific spanning arborescence of the conservation graph, and consequently recover the original network topology from it. 

\begin{figure}[H]
\centering
\includegraphics[scale=0.5]{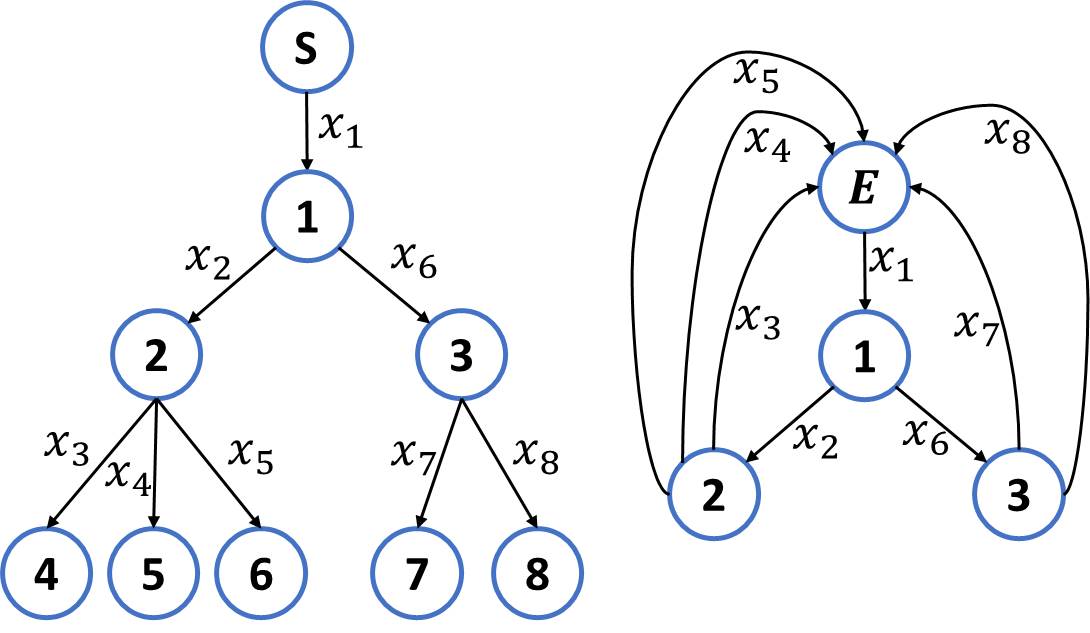}
\caption{Topology of an arborescence network and its conservation graph}
\label{Fig:Radial_Network}
\end{figure}

It may be noted that the conservation graph of an arborescence network has multiple spanning arborescences and the specific spanning arborescence rooted at the environment node enables reconstruction of the original network. For example, consider an arborescence network and its conservation graph $G_c$ shown in Fig.~\ref{Fig:Radial_Network}.  A spanning arboroscence of the conservation graph rooted at node $E$ formed by non-sink flow edges $x_1,x_2,x_6$ as branches, and sink flow edges $x_3,x_4,x_5,x_7,x_8$ as chords is shown in Fig ~\ref{Fig:Arb_Realisation}. From this specific spanning arborescence, we can obtain the original network topology by removing the environment node and connecting outgoing edge $x_1$ to source node and incoming edges $x_3, x_4, x_5, x_7, x_8$ to individual sink nodes. On the other hand, from the spanning arborescence of the conservation graph rooted at node $3$ shown in Fig. \ref{Fig:Arb_Realisation}, it is not possible to reconstruct the original network.  We denote the spanning arborescence of $G_c$ rooted at node $E$ as $S_E$ and the f-cutset matrix corresponding to $S_E$ as $\Cbf_{S_E}$. The following section describes our proposed procedure to obtain $\Cbf_{S_E}$ from flow measurements and the reconstruction of the arborescence network topology.

\section{Network Reconstruction from Flows} \label{Sec:Method}
The input data are measurements $\xbf_i$ of flows in all $e$ edges at $n_s (> e)$ different steady states. The $n_s$ samples of $\xbf_i$ are stacked into a data matrix $\Xbf_{e \times n_s}$ where each column of the data matrix is a sample. We first describe how to identify an f-cutset matrix of the conservation graph from perfect measurements which do not contain any measurement noise, before extending the approach to deal with noisy measurements. 
\subsection{Identifying an f-cutset matrix from data}
The conservation equations (Eq. \ref{Eq:conservation}) imply that the rows of $\Abf_{ir}$ are orthogonal to the  samples $\xbf_i$. In other words, these samples lie in the null space of $\Abf_{ir}$. We use the \emph{singular value decomposition} (SVD) to obtain a basis for the subspace which is orthogonal to $\Xbf$, because, as demonstrated later, it can be used even if measurements are noisy. The SVD of the data matrix is given by
\begin{equation}
\Xbf = \Ubf\Sbf\Vbf = \Ubf_1\Sbf_1\Vbf_1^T + \Ubf_2\Sbf_2\Vbf_2^T
\end{equation}
where $\Ubf_1$ and $\Vbf_1$ are the left and right singular vectors corresponding to the non-zero singular values of $\Xbf$, and $\Ubf_2$ and $\Vbf_2$ are the right singular vectors corresponding to the \emph{zero} singular values of $\Xbf$. Since the rank of $\Abf_{ir}$ is $m$, $\Xbf$ will have $m$ zero singular values and $(e-m)$ non-zero singular values. Using the fact that the singular vectors are orthonormal and the matrix $\Sbf_2$ is a matrix of zeros it follows that 
\begin{align}
\Ubf_2^T\Xbf &= \Sbf_2\Vbf_2 = \mathbf{0} \label{Eq:svd}
\end{align}
\eqref{Eq:svd} implies that the columns of $\Ubf_2$ form a basis for the subspace orthogonal to columns of $\Xbf$. Consequently the row space of $\Ubf_2^T$ and the reduced incidence matrix $\Abf_{ir}$ are identical.  Therefore
\begin{align}\label{Eq:rotation}
\Ubf_2^T = \Mbf\Abf_{ir}
\end{align}
where $\Mbf : m \times m$ is some non-singular rotation matrix.  

We define a valid partition of the variables $\xbf$ into a set of $m$ dependent variables $\xbf_D$ and remaining independent variables $\xbf_I$ such that the $m \times m$ sub-matrix $\Abf_D$ of $\Abf_{ir}$ corresponding to columns of $\xbf_D$ is non-singular.  The conservation equations can then be equivalently written as
\begin{align} \label{Eq:Implicit}
\Abf_D^{-1}\begin{bmatrix}
\Abf_D & \Abf_I 
\end{bmatrix}
\begin{bmatrix}
\xbf_D \\ \xbf_I 
\end{bmatrix} = 
\begin{bmatrix}
\Ibf_m & \Abf_D^{-1}\Abf_D 
\end{bmatrix}
\begin{bmatrix}
\xbf_D \\ \xbf_I 
\end{bmatrix} \nonumber \\ = 
\begin{bmatrix}
\Ibf_m & \Rbf_D 
\end{bmatrix}
\begin{bmatrix}
\xbf_D \\ \xbf_I 
\end{bmatrix} = \mathbf{0}.
\end{align}
In Theorem~\ref{Thm:Partition}, we prove that corresponding to every valid partition the matrix $\Rbf_D$ is unique.  We also prove that even if we start with an estimate $\Ubf_2^T$ derived from data (which is related to $\Abf_{ir}$ through Eq. \ref{Eq:rotation}), corresponding to every valid partition of the variables the matrix $\Rbf_D$ is exactly obtained.
In Theorem~\ref{Thm:CutsetMatrix}, we prove that the matrix $[\Ibf_m \ \ \Rbf_D]$ is an f-cutset matrix corresponding to some spanning tree of the conservation graph.  The edges corresponding to $\xbf_D$ are branches and the remaining edges are chords of the spanning tree.

\begin{figure}[H]
		\centering
		\includegraphics[scale=0.4]{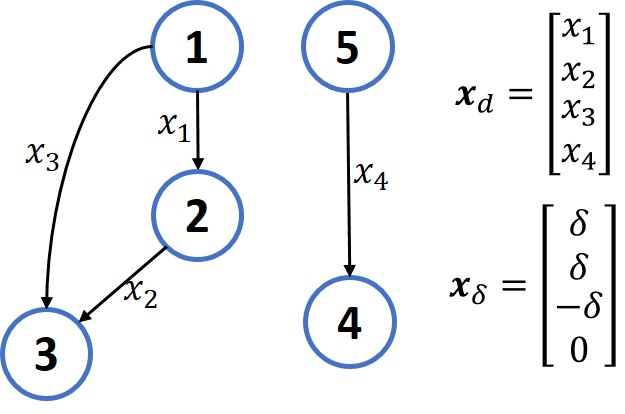}
		\caption{Theorem~\ref{Thm:CutsetMatrix}:Graph for illustration}
		\label{Fig:Thm_Example}
	\end{figure}

\begin{Theorem} \label{Thm:Partition}
	Given a linear model $\Abf\xbf=\mathbf{0}$, there exists a unique matrix $\Rbf_D = \Abf_D^{-1} \Abf_I$ with respect to any valid partition of the variables $\xbf = [\xbf_D \;|\; \xbf_I]$ such that $\xbf_D = -\Rbf_D\xbf_I$.
\end{Theorem}
{\bf Proof.}
Let us consider a valid partition of the variables $\xbf = [\xbf_D \;|\; \xbf_I]$ which implies that $\Abf_D$ is non-singular. Therefore,
\begin{align*}
\Abf \xbf &= \Abf_D \xbf_D + \Abf_I \xbf_I = \mathbf{0} \\
\implies \xbf_D &= -\Abf_d^{-1} \Abf_I \xbf_I = \Rbf_D \xbf_I
\end{align*}
Now, we rotate $\Abf$ with a non-singular matrix $\Mbf$ and repeat the partitioning as follows:
\begin{align*}
\Mbf \Abf \xbf &= \Mbf \Abf_D \xbf_D + \Mbf \Abf_I \xbf_I = \mathbf{0} \\
\implies \xbf_D &= -(\Mbf \Abf_D)^{-1} (\Mbf \Abf_I) \xbf_I \\
\implies \xbf_D &= -\Abf_D^{-1} (\Mbf^{-1} \Mbf) \Abf_I \xbf_I = \Rbf \xbf_I
\end{align*}
Therefore, $\Rbf_D$ is unique for any valid partition of $\xbf$ irrespective of the model matrix $\Abf$ used to obtain it. {\bf Q.E.D}
\begin{Theorem} \label{Thm:CutsetMatrix}
	The matrix $\begin{bmatrix} \Ibf_m & \Rbf_D \end{bmatrix}$ corresponding to a valid partition of $\xbf = [\xbf_D \;|\; \xbf_I]$, is the f-cutset matrix $\Cbf_{x_D}$ of $G_c$ with respect to the spanning tree formed by edges $\xbf_D$. 
\end{Theorem}
{\bf Proof.}
	We prove this by contradiction assuming edges $\xbf_D$ do not form a spanning tree. Suppose the edges in $\xbf_D$ do not form a spanning tree, then they must form a loop. For example, refer to graph in Fig.~\ref{Fig:Thm_Example} with 5 nodes and 4 edges but not forming a spanning tree. Any arbitrary value $\delta$ can be added to the flows in a loop with appropriate signs such that conservation is not affected as shown in Fig.~\ref{Fig:Thm_Example}. We know that $\Abf_D\xbf_D = \Abf_I\xbf_I$ and as some edges in $\xbf_D$ form a loop, we can write $\Abf_D(\xbf_D+\xbf_\delta) = \Abf_I\xbf_I$, where $\xbf_\delta$ is a vector comprising of non-zero elements of equal magnitude $\delta$ and appropriate sign corresponding to the edges in $\xbf_D$ forming a loop. This implies that $\Abf_D$ is not invertible because $\xbf_\delta$ can take many values satisfying the equation. This is a contradiction because we started with a valid partition and $\Abf_D$ is invertible. Hence the edges in $\xbf_D$ form a spanning tree. 
	
	Now, as per theorem~\ref{Thm:Partition}, $\Rbf_D$ is unique for a particular combination of $\xbf_D$ and $\xbf_I$. Since $\begin{bmatrix} \Ibf_m & \Rbf_D \end{bmatrix}$ has the form of a cutset matrix and $\Rbf_D$ is unique, $\begin{bmatrix} \Ibf_m & \Rbf \end{bmatrix}$ is an f-cutset matrix of $G_c$.
{\bf Q.E.D}

Exploiting the above two theorems, we can learn an f-cutset matrix corresponding to some spanning tree of the conservation graph using the following two steps.
\begin{itemize}
    \item Apply SVD to the data matrix to obtain a basis for subspace orthogonal to $\Xbf$ i.e., $\Ubf_2^T$.
    \item Determine a valid partition of the variables and transform $\Ubf^T_2$ to the form $[\Ibf_m \ \ \Rbf_D]$ to obtain an f-cutset matrix.
\end{itemize}
\subsection{Obtaining the f-cutset matrix $\Cbf_{S_E}$} \label{Sec:f-cutset}
In the preceding section, we have shown that it is possible to obtain an f-cutset matrix corresponding to some spanning tree of the conservation graph.  We now develop a procedure for transforming this f-cutset matrix to the specific f-cutset matrix $\Cbf_{S_E}$. The following theorem is utilized for this purpose.
\begin{Theorem} \label{Thm:Radial1}
Any flow conservation equation corresponding to a cutset of the conservation graph of an arborescence network, contains exactly one flow with a coefficient of $+1$ or $-1$, while all other flows in the equation have coefficients with opposite sign.  The flow with the unique coefficient corresponds to the lowest level non-sink flow.
\end{Theorem}
{\bf Proof.}
Every cutset eq. must contain a non-sink flow because removal of only sink flow edges will not cut the conservation graph into 2 components. Therefore, a cutset eq. contains one or more non-sink flows (non-leaf edges). Let the non-leaf edges in the cutset be at levels $l_1,l_2,\ldots$ such that $l_1<l_2<\ldots$. 

Firstly, we show that there exists only one non-leaf edge at the smallest level $l_1$. Suppose there exist two non-leaf edges $x_j,x_k$ at level $l_1$. These edges $x_j$ and $x_k$ are only connected via their ancestor edges at levels smaller than $l_1$. Since there are no edges at levels smaller than $l_1$ in the cutset, $x_j$ and $x_k$ remain independent which is a contradiction because they exist in the same cutset eq. as per the assumption. Hence, $x_j$ and $x_k$ at smallest level $l_1$ cannot be in the same cutset.  

Secondly, consider the case when there is only one edge $x_j$ at level $l_1$. The edges at levels greater than $l_1$ can be categorised at descendant and non-descendant edges to edge $x_j$. Since edge $x_j$ is connected to these non-descendant edges only via its ancestor edges and other edges at level $l$, without them no dependency exists between the flow $x_j$ and flows along the non-descendant edges. Therefore, every cutset in the conservation graph of an arborescence network comprises of a non-leaf edge $x_j$ and only its descendant edges. In such a case, flow $x_j$ sums up the flows in the descendant edges. Therefore $x_j$ have the unique coefficient in the cutset eq. and it belongs to the level level $l_1$ non-sink flow among the edges in the cutset. 
{\bf Q.E.D}
 
Starting with the first row of the f-cutset matrix $\Cbf_{x_D}$, we determine the edge with the unique coefficient.  If this edge is a branch, we proceed with the next row of $\Cbf_{x_D}$.  If this edge is a chord, then we interchange it with the corresponding branch edge of this f-cutset.  This chord-branch interchange is also known as the \emph{elementary tree transformation technique} \cite{Deo} and results in a new spanning tree of the conservation graph.  The f-cutset matrix corresponding to this new spanning tree is found by eliminating the new branch edge from all other f-cutsets in which it may be present.  The coefficient corresponding to this new branch is also made positive by reversing the signs of coefficients of all edges in this cutset.  At the end of this intercharge, by Theorem~\ref{Thm:Radial1}, it is guaranteed that the new branch is a non-sink flow edge.  At the end of this procedure, we obtain $\Cbf_{S_E}$ in which all non-sink flow edges are branches, and the corresponding spanning arborescence is $S_E$.  Algorithm \ref{Algo1} summarizes the steps of this procedure.

\begin{algorithm} 
\caption{Algorithm to obtain $\Cbf_{S_E}$} \label{Algo1}
\begin{algorithmic} 
\State \textbf{Input:} $\Cbf_{x_D} = \begin{bmatrix}
\Ibf_m & \Rbf_D \end{bmatrix}$
\State \textbf{Output:} $\Cbf_{S_E}$
\State Let $C_{kj}$ be the an element $\Cbf_{x_D}$ in $k^{th}$ row and $j^{th}$ column 
\For {$k$ in 1 to $m$}
\If {$C_{kj} \leq 0 \;\; \forall \; j>m$}
  	\State \Continue
\Else
    \State $l= \{j: C_{kj} = -1\}$
    \State Interchange the columns $k$ and $l$ of $\Cbf_{x_D}$. 
    \State Perform necessary row operations to transform the first $m$ columns into an identify matrix to obtain the updated f-cutset matrix $\Cbf_D$
\EndIf
\EndFor
\end{algorithmic}
\end{algorithm}

\begin{remark}
In Algorithm~\ref{Algo1}, the lowest level non-sink flow in a f-cutset can also be identified as the edge corresponding to maximum flow magnitude among all the edges. However, this approach may occasionally fail when dealing with noisy data.
\end{remark}
\subsection{Reconstructing arborescence network} \label{Sec:Topo_ID} 
In order to construct $S_E$ given $\Cbf_{S_E}$, we make use of the following two theorems:
\begin{Theorem} \label{Thm:fcutsetchords}
Each f-cutset of $\Cbf_{S_E}$ corresponding to a non-sink flow edge as its branch contains all its descendant sink flow edges as chords and no other edges. \end{Theorem}
The following corollaries can also be derived from the above theorem.
{\bf Proof.}
Firstly, every cutset in $\Cbf_{S_E}$ contains only non-sink flow edges as branches and sink flow edges as chords. The structure of $\Cbf_{S_E}$ implies that in each cutset equation, a non-sink flow sums up a combination of sink flows. Now, a non-sink edge is connected to non-descendant sink flows only via other non-sink flows which do not appear in the same cutset equation. Therefore in every cutset of $\Cbf_{S_E}$, the branch corresponding to a non-sink flow edge contains only its descendant sink flow edges as chords. 
{\bf Q.E.D}

\begin{corollary}\label{fcutsetchordsCor1}
    Two f-cutsets of $\Cbf_{S_E}$ contain one or more common sink flow edges as chords if and only if the non-sink flow edge of one cutset is a descendent of the non-sink flow edge of the other cutset. 
\end{corollary} 
\begin{corollary}\label{fcutsetchordsCor2}
    If non-sink flow edge $x_j$ is a descendant of non-sink flow edge $x_k$, then the chords of f-cutset of $\Cbf_{S_E}$ corresponding to $x_j$ will be a subset of the chords of f-cutset of $\Cbf_{S_E}$ corresponding to $x_k$. 
\end{corollary}
Corollary \ref{fcutsetchordsCor1} implies that if a set of f-cutsets have common chords, then the corresponding branches form a dipath. Corollary \ref{fcutsetchordsCor2} implies that among these cutsets, the branch of the cutset with fewer chords will be at a higher level in the dipath.  Based on this inference, Algorithm~\ref{Algo2} derives the arborescence network topology by first determining the connectivity of non-sink flow edges, followed by the connectivity of sink flow edges. In Algorithm~\ref{Algo2}, we firstly we re-arrange the rows of $\Cbf_{S_E}$ in the descending order of number of non-zero elements in the rows. Since the data set only labels the $e$ edges, we follow a specific convention in labelling the nodes of the network. As all the nodes other than the source have exactly one incoming edge, these nodes are given the same labels as the incoming edges. The source node is given the label $e+1$. This convention is followed in Algorithm~\ref{Algo2} while reconstructing the network. Arrays $\sbf$ and $\tbf$ contain the source and destination nodes of the directed edges of the network, while the array $\xbf_e$ contains the labels of edges corresponding to dependent and independent variables in the order of columns of $\Cbf_{S_E}$.

\begin{algorithm} 
\caption{Realization of Topology Graph from $\Cbf_{S_E}$} \label{Algo2}
\begin{algorithmic}
\State \textbf{Input:} $\Cbf_{S_E}$
\State \textbf{Output:} Arrays $\sbf$ and $\tbf$
\State Initialize $\sbf$, $\tbf$ to be empty arrays 
\State Sort rows of $C_{S_E}$ in the descending order of number of non-zero elements
\State Re-arrange the first $m$ columns of $C_{S_E}$ to obtain identity matrix and re-order $\xbf_e$ accordingly
\State $\sbf[1] \gets (e+1) \; ; \; \tbf[1] \gets \xbf_e[1]$
\For{$k$ in 2 to $m$}
\State Set $r_1 = \{j: C_{S_E,kj} \neq 0 \}$
\For {$p$ in $k-1$ to 1}
\State Set $r_2 = \{j: C_{S_E,pj} \neq 0 \}$
\If{$r_1 \cap r_2 \neq \phi $}
  	\State $\sbf[k] \gets p \; ; \; \tbf[k] \gets x_e[k]$
\EndIf
\EndFor
\EndFor
\For{$j$ in $(m+1)$ to $e$}
 \State Set $c = \{k: C_{S_E,kj} = -1 \}$
 \State $\sbf[j] \gets \xbf_e[\max \; c] \; ; \; \tbf[j] \gets \xbf_e[j]$
 \EndFor
\end{algorithmic}
\end{algorithm}

\section{Illustrative Example}
Consider the arborescence network shown in Fig.~\ref{Fig:Radial_Network} for which conservation equations can be written around nodes $1$, $2$, and $3$. We generate 10 samples of all the flows in the network satisfying the conservation equations by choosing random samples for $x_3,x_4,x_5,x_7$ and $x_8$ from a normal distribution with a mean of $10$ and a variance of $4$. The flows of $x_1$, $x_2$, and $x_3$ are computed using the conservation equations. The data matrix is provided in Table~\ref{Tab:Data}.

\begin{table}[h] 
\centering
\caption{Data Matrix $\Xbf$}
\begin{tabular}{|l|l|l|l|l|l|l|l|l|l|} 
\hline
49.697 & 48.544 & 50.089 & 51.141 & 51.021 & 50.295 & 46.709 & 57.293 & 48.351 & 54.665 \\ \hline
31.221 & 30.750 & 28.302 & 32.230 & 28.874 & 27.453 & 28.401 & 32.244 & 29.580 & 33.775 \\ \hline
8.327  & 9.849  & 7.587  & 13.217 & 8.144  & 6.638  & 8.898  & 12.890 & 10.402 & 13.158 \\ \hline
10.899 & 10.413 & 10.858 & 10.431 & 10.385 & 11.147 & 10.298 & 10.437 & 9.114  & 12.606 \\ \hline
11.995 & 10.487 & 9.857  & 8.582  & 10.345 & 9.668  & 9.205  & 8.917  & 10.063 & 8.011  \\ \hline
18.476 & 17.794 & 21.787 & 18.911 & 22.147 & 22.842 & 18.307 & 25.049 & 18.771 & 20.890 \\ \hline
8.156  & 9.529  & 13.673 & 10.566 & 10.904 & 9.733  & 7.306  & 12.143 & 10.319 & 12.834 \\ \hline
10.320 & 8.266  & 8.114  & 8.345  & 11.243 & 13.109 & 11.001 & 12.906 & 8.452  & 8.056  \\ \hline
\end{tabular} \label{Tab:Data}
\end{table}

\subsection{Obtaining an f-cutset matrix from flow data}
Applying SVD to the data matrix the following singular values are obtained.
\begin{align*}
\begin{bmatrix}
210.87 & 9.62 & 6.99 & 5.36 & 2.63 &  0 &  0 &  0 \end{bmatrix}
\end{align*}
Based on the number of zero singular values, the dimension of the null space of the data matrix is correctly estimated as $3$.  From the transpose of the right singular vectors corresponding to the zero singular values, an estimate of the linear conservation equations is obtained as (rounded to 2 decimals):
\begin{align*}
& \Abf =\Ubf_2^T = \\
  &  \begin{bmatrix}
    0.63 & -0.55 & -0.08 & -0.08 &	-0.08 &	-0.33 &	-0.30 & -0.30 \\
0.02 & 0.39 & -0.40 & -0.40 & -0.40 & 0.34 & -0.36 & -0.36 \\
0.14 &	0.19 & -0.33 & -0.33 & -0.33 & -0.54 & 0.40 & 0.40
    \end{bmatrix}
\end{align*}

\begin{figure}[H]
\centering
\includegraphics[scale=0.5]{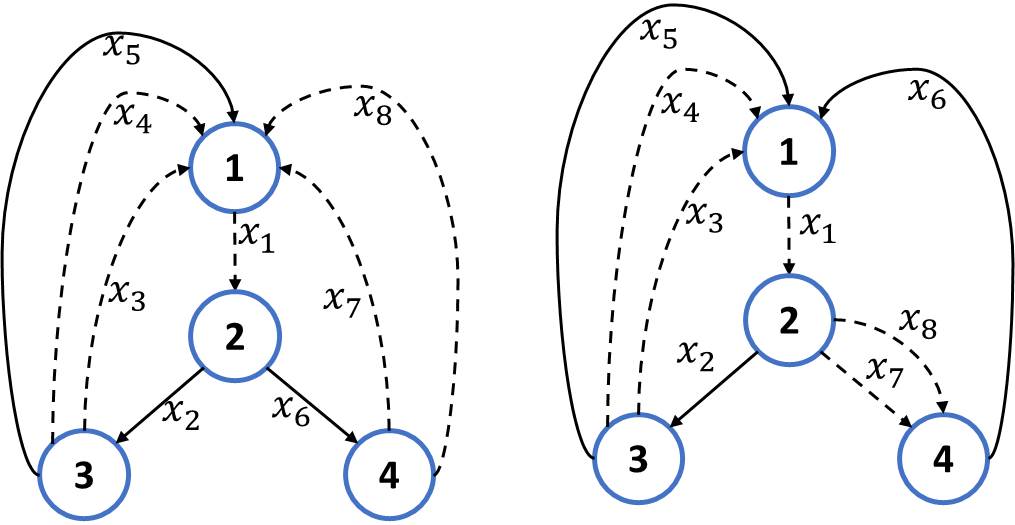}
\caption{2-Isomorphic graphs realised from f-cutset matrix}
\label{Fig:Realisation}
\end{figure}

The set $x_D = [x_2, x_5, x_6]$ can be chosen as a valid partition, since the corresponding sub-matrix of $\Abf$ is non-singular.  The f-cutset matrix corresponding to this partition is obtained from $\Abf$ as: 
\begin{align*}
 \Cbf_{x_D} = \begin{array}{c c}  &
\begin{array}{c c c c c c c c c} x_2 & x_5 & x_6 \;\; & x_1 & x_3 & x_4 & x_7 & x_8 \\
\end{array} 
\\
&
\left[
\begin{array}{c c c c c c c c}
1 & \;\; 0 &  \;\; 0 & \;\; -1 & \;\; 0 & 0 & 1 & 1 \\
0 & \;\; 1 &  \;\; 0 & \;\; -1 & \;\; 1 & 1 & 1 & 1 \\
0 &	\;\; 0 &  \;\; 1 & \;\; 0 & \;\; 0 & 0 & -1 & -1
\end{array}
\right] 
\end{array}
\end{align*}
This is a f-cutset matrix with edges $x_2,x_5,x_6$ forming a spanning tree but it is not in the desired one. Consequently, graph realisation may not give a graph identical to the original due to 2-isomorphism. In Fig.~\ref{Fig:Realisation}, we show two possible graph realisations from the obtained f-cutset matrix. One is identical to the original $G_c$ while the other is a 2-isomorhpic graph. 

To get the desired f-cutset matrix, we apply the steps in Algorithm~\ref{Algo1} and get:
\begin{align*}
\Cbf_{S_E} = \begin{array}{c c}  &
\begin{array}{c c c c c c c c c} x_1 \;\; & x_2 \; & x_6 \;\; & x_5 \;\; & x_3 \;\; & x_4 \;\; & x_7 \;& x_8 \\
\end{array} 
\\
&
\left[
\begin{array}{c c c c c c c c}
\; 1 & \;\; 0 & \;\; 0 & -1 & -1 & -1 & -1 & -1 \\
\; 0 &	\;\; 1 & \;\; 0 & -1 & -1 & -1 & 0 & 0 \\
\; 0 & \;\; 0 & \;\; 1 & 0 & 0 & 0 & -1 & -1 
\end{array}
\right] 
\end{array}
\end{align*}
with $(x_1,x_2,x_6)$ as the branches and $(x_5,x_3,x_4,x_7,x_8)$ as the chords. 

\begin{figure}[H]
\centering
\includegraphics[scale=0.5]{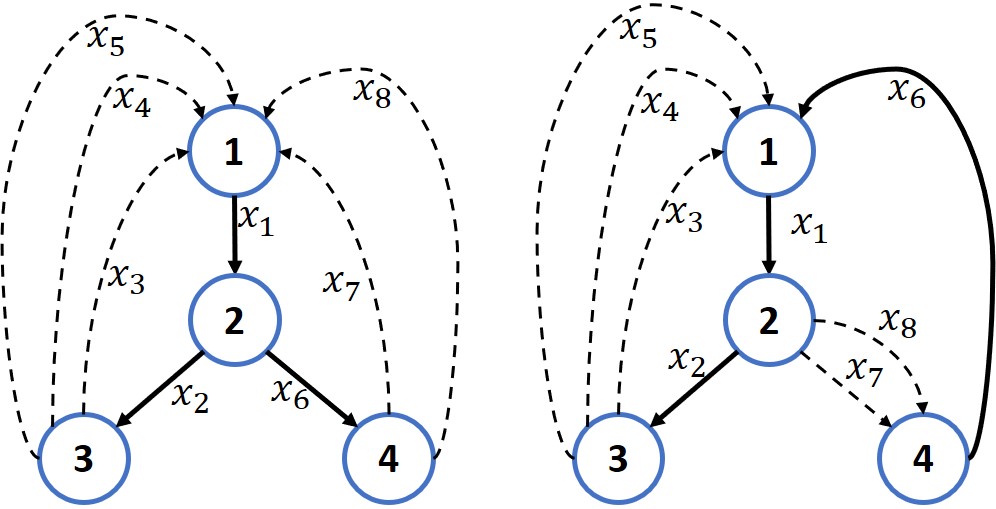}
\caption{2-Isomorphic graphs with spanning arborescence}
\label{Fig:Arb_Realisation}
\end{figure} 

\subsection{Topology Identification}
Though the desired f-cutset matrix is obtained, any graph or arborescence realisation algorithm might still result in a 2-isomorphic graph. Fig.~\ref{Fig:Arb_Realisation} shows two possible 2-isomorphic graph realisations of desired f-cutset matrix. To overcome this, Algorithm~\ref{Algo2} is applied on $\Cbf_c$ to realise the topology graph. It can be observed that $x_1$ is the source flow since it is written as a linear combination of all sink flows $(x_3,x_4,x_5,x_7,x_8)$. $x_2$ and $x_6$ are connected to the terminal node of $x_1$ since they have common edges with $x_1$ in their linear combinations. $x_4$ is connected to the dipath $x_1,x_2$ as indicated by the corresponding column in $\Cbf_f$ and similarly other sink flow edges are connected. Finally, we get the original topology shown in Fig.~\ref{Fig:Radial_Network}.

\section{Time Complexity}
In this section, we analyse the time complexity of the proposed methodology for arborescence networks. This assumes significance because in \cite{arya2013}, it is discussed that reconstructing the topology of a radial power distribution network is NP-hard when posed as an optimization problem. Since the objective is to show that the overall algorithm runs in polynomial time complexity, we restrict the analysis to the ideal case of noise-free data.  The first step is applying SVD on $\Xbf_{e \times n_s}$ which is shown to be of complexity $\mathcal{O}(e^2n_s+n_s^3)$ \cite{Golub}. The next step involved computing RREF of a matrix of dimension $(m \times e)$ which is of complexity $\mathcal{O}(m^2e)$ \cite{fraleigh95}. Coming to Algorithm~\ref{Algo1}, it involves a loop with $m$ iterations involving $m$ RREF operations. This step involves a complexity of $\mathcal{O}(m \times (e+m^2e))$. The final step of topology identification involves graph realization for which we proposed Algorithm~\ref{Algo2} which is of complexity $\mathcal{O}(m^2)$. Since all the steps involved can be performed in polynomial time, the overall algorithm is of polynomial time complexity. 

\section{Noisy Measurements} \label{Sec:Noisy}
In general, the measurements are error-prone due to noise and we can model them as follows:
\begin{align}
\ybf_j &= \xbf_j + \epsilonbf_j \\
\Ybf &= \Xbf + \Ebf,
\end{align}
where $\epsilonbf_j$ is the random error vector at instance $j$ and $\Ebf$ is the matrix of errors. It is assumed that the PCA relates assumptions are satisfied by these measurements \cite{Naras15}. In this case, the dimension ($m$) of the desired subspace $\mathcal{S}$ is not given by the number of zero singular values. To be able to identify $\mathcal{S}$ from $\Ybf$, we propose to use PCA and its variants which work under different assumptions of $\mubf$ and $\Sigmabf_e$. The linear model given by PCA is the an approximation of the true model and denoted by $\hat \Abf$. While applying PCA, we use the fact that the sample covariance matrix $\Sbf_y = \frac{1}{n_s}\Ybf\Ybf^T$ will have $(n-m)$ dominant eigen values and $m$ small eigen values, under the assumption that SNR is high. The following are different cases of noise distribution under which PCA can identify the $\mathcal{S}$ in which the noise-free data lies:
\subsection{Homoscedastic Noise} 
In this case, it is assumed that noise is i.i.d. i,e,. $\mubf = \mathbf{0} ; \Sigmabf_e = \sigma^2 \Ibf$. In this case, the smallest $m$ eigen values of the expected covariance matrix will be equal to $\sigma^2$. This fact can be used to identify the dimension of $\mathcal{S}$. Since we only know the sample covariance matrix $\Sbf_y$, we have to perform a hypothesis test to verify how many of the smallest eigen values of the covariance matrix are actually equal, to ascertain the true value of $m$ \cite{Jolliffe}. Further, the values in non-identity part of $\text{rref}(\hat \Abf)$ will not be exactly $+1,-1$ or $0$ and cannot be interpreted graphically. Hence, each of these values are rounded to their nearest neighbour among $+1,-1$ and $0$. 

\subsection{Heteroscedastic Noise} In this case, it is assumed that the noise variables are independent but not identical and $\Sigmabf_e$ is known. For this case, in \cite{Naras08}, it has been shown that by pre-multiplying $\Ybf$ with the inverse of Cholesky factor of $\Sigmabf_e$, the transformed data matrix ($\Ybf_s$) will have i.i.d. errors. Further, the smallest $m$ eigen values of data covariance matrix will be equal to one. Thus, $\Ybf_s$ can be used to determine the value of $m$ and a basis for $\mathcal{S}$ after performing hypothesis testing as in previous case. It is to be noted that even in case of homoscedastic noise, this transformation can be applied to obtain $m$ smallest eigen values equal to one. This enables us to apply the same methodology in both the cases.


\subsection{Methodology}
In general, the following are the steps to identify the topology of a arborescence network from noisy measurement matrix $\Ybf$ and known $\Sigmabf_e$:
\begin{enumerate}
\item Find the Cholesky decomposition of $\Sigmabf_e$ given by: $\Sigmabf_e = \Lbf\Lbf^T$
\item Transform the data matrix to get: $\Ybf_s = \Lbf^{-1}\Ybf$
\item Apply PCA on $\Ybf_s$: $\text{SVD}(\frac{\Ybf_s}{\sqrt{n_s}}) = \Ubf_s\Sbf_s\Vbf_s^T$
\item For different values $\geq 2$, perform hypothesis testing to establish the true value of $m$. 
\item After establishing the value of $m$, choose the last $m$ eigen vectors of sample covariance matrix of $\Ybf_s$ and transform them to get:
$\hat \Abf = \Ubf_{2s}^T\Lbf^{-1}$
\item Find the RREF of $\hat \Abf$ obtained. 
\item Round each of the elements of non-identity part of $\text{rref}(\hat \Abf)$ to the nearest among $-1 , 0$ and $+1$.
\item Apply Algorithm~\ref{Algo1} to obtain f-cutset matrix in desired form.
\item Finally, apply the graph realization algorithm to obtain the topology.
\end{enumerate}

In the following section, we apply this methodology to identify the topology of a simple arborescence network from noisy measurements. 

\section{Results and Discussions}
\begin{figure*}
    \centering
    \includegraphics[scale=0.8]{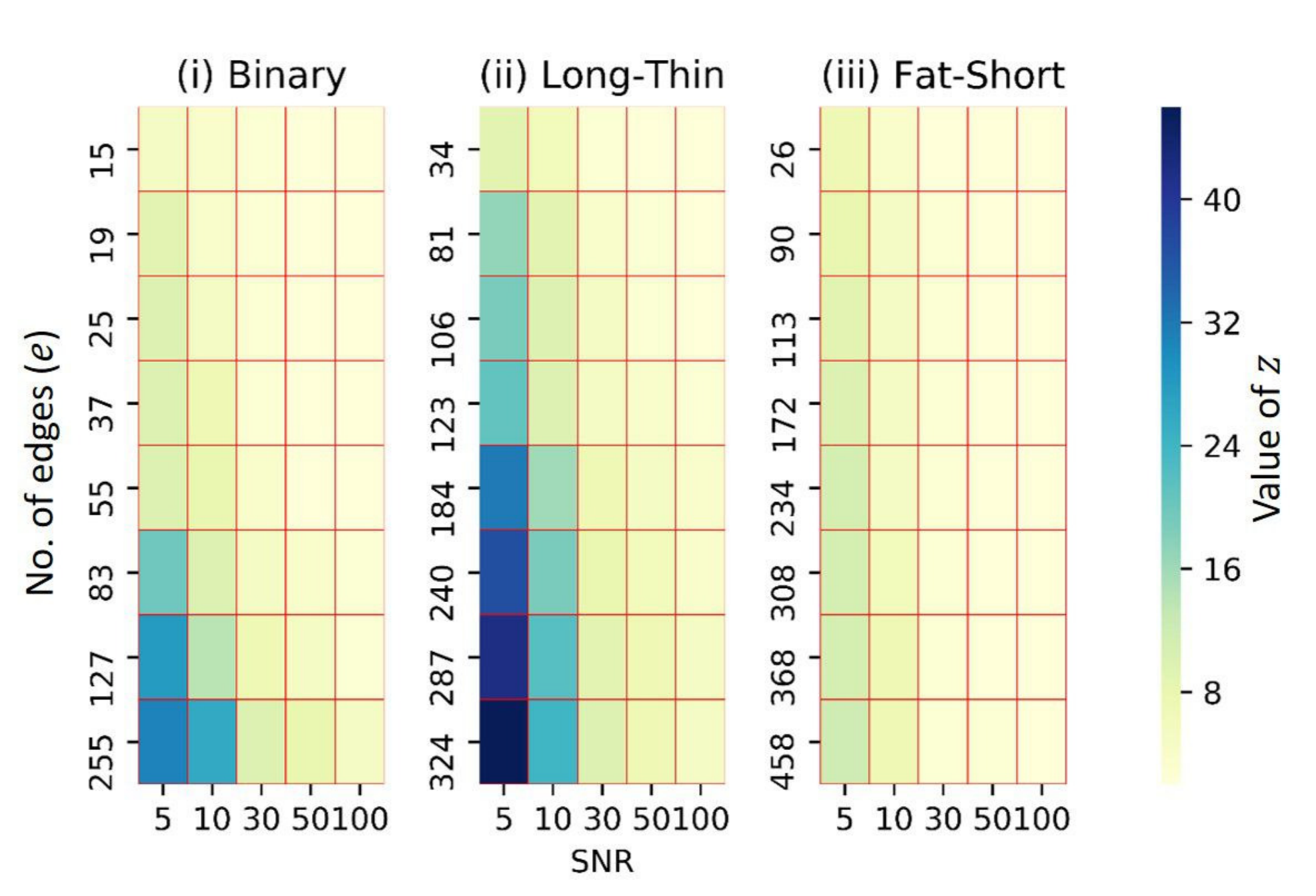}
    \caption{Simulations results: No. of samples required for $100\%$ accurate network reconstruction}
    \label{fig:plots}
\end{figure*}

To corroborate the proposed method, we apply it on randomly simulated arborescence networks and data. We categorise the arborescene networks into three types to study the effect of the tree structure on topology identification. The types include \textit{Binary networks} - networks with tree topology in which every parent node has exactly two child nodes, \textit{Thin-Long trees} - networks with many tree layers but fewer nodes at each layer, and \textit{Fat-Short trees} - networks with fewer tree layers but many nodes at each layer. The simulations are conducted on MATLAB 2018a as follows:
\begin{enumerate}
\item For each type of conservation network, 8 different networks were randomly generated beginning with a root node and then randomly assigning number of layers and number of child nodes to each parent node. These random numbers were chosen from a range decided based on the type of the network. As a result, each network has a different number of edges as shown in Fig.~\ref{fig:plots}.
\item Then, for each network, a dataset of $n_s$ samples per flow variable are randomly generated as follows:
\begin{enumerate}
\item Firstly, the samples for sink flows are samples randomly from one of the three Gaussian distributions with means $(100,200,300)$ and standard deviations $(10,20,30)$, respectively.  
\item For each of the non-sink flows, the sample values are determined by summation of sink flows incident on nodes which are descendants to the nodes on which the non-sink flows are incident.
\end{enumerate}
\item Then, i.i.d. noise is added to each of the samples based on a chosen SNR value.
\end{enumerate}
This simulation is repeated 100 times for each network and with different SNR values of (100,50,30,10,5) and the numbers of samples $n_s$ required for $100\%$ accurate identification are obtained. Here, the number of samples are varied as $z$ multiples of $e$ i.e., $N_s = ze$. These results are plotted in a heatmap in the Fig.~\ref{fig:plots}. The colours in the heatmap indicate the value of $z$ in a range of 0-50 as shown in the colour scale.   

To make conclusions from the heatmap, consider the binary network with $e=255$, long-thin network with $e=240$ and fat-short network with $e=234$ as the number of edges are comparable. It can be observed that the colours in the heatmap for these networks at SNR=5 are widely different indicating that the binary network and the long-thin network require significantly more number of samples than the fat-short network. In general, it can be concluded that for a given SNR and given number of edges in a comparable range, the binary networks and long-thin networks require more samples than short-fat networks.  This can be attributed to two causes. For the similar number of nodes (i) binary networks and long-thin networks have more number of linearly independent equations of conservation compared to short-fat networks and (ii) the cutset matrix $\Cbf_f$ is more sparse for binary networks and long-thin networks.

\section{Conclusions}
Reconstruction of network topology is an important problem in network science. This work deals a network reconstruction problem related to  conserved networks arising in engineering, biology, and financial fields. A novel concept of conserved graphs is introduced to analyse conserved networks. Further, properties of conservation graphs are discussed from a network reconstruction viewpoint. Then, these properties have been exploited to reconstruct network using principal component analysis. A novel algorithm  to reconstruct conserved networks from steady-state flow data has been proposed. The reconstruction algorithm has been corroborated with three different kinds of conserved networks arising in different fields. A comparison study in terms of signal to noise ratio in data, number of nodes, measurement required  for reconstructing 100\% accurate network has been performed for all three types of conserved networks. 




\acknow{Please include your acknowledgments here, set in a single paragraph. Please do not include any acknowledgments in the Supporting Information, or anywhere else in the manuscript.}

\showacknow{} 


\bibliography{pnas-ref}

\end{document}